\documentclass[11pt,a4paper]{article}
\usepackage{graphicx} 
\usepackage{authblk}
\usepackage{geometry}
\usepackage{tabularx}
\usepackage{subcaption}

\usepackage{amssymb}
\usepackage{rotating}
\usepackage{amsmath}
\usepackage{bm}
\usepackage{url}

\title{Trade-offs Between Individual and Group Fairness in Machine Learning: A Comprehensive Review} 

\author[1,2]{Sandra Benítez-Peña}
\author[2]{Blas Kolic}
\author[2]{Victoria Menendez}
\author[2,3,$\dagger$]{Belén Pulido}

\affil[1]{{Departamento de Estadística, Universidad Carlos III de Madrid, Getafe, Spain}}
\affil[2]{{uc3m-Santander Big Data Institute (IBiDat), Universidad Carlos III de Madrid, Getafe, Spain}}
\affil[3]{{Department of Statistics and O.R., Universidad Nacional de Educación a Distancia (UNED), Madrid, Spain }}
\affil[$\dagger$]{\textit{belen.pulido@ccia.uned.es}}

\usepackage[dvipsnames]{xcolor}

\usepackage{pifont}
\newcommand{\xmark}{\ding{55}} 

\begin{document}

\maketitle

\begin{abstract}
Algorithmic fairness has become a central concern in computational decision-making systems, where ensuring equitable outcomes is essential for both ethical and legal reasons. Two dominant notions of fairness have emerged in the literature: Group Fairness (GF), which focuses on mitigating disparities across demographic subpopulations, and Individual Fairness (IF), which emphasizes consistent treatment of similar individuals. 
These notions have traditionally been studied in isolation. In contrast, this survey examines methods that jointly address GF and IF, integrating both perspectives within unified frameworks and explicitly characterizing the trade-offs between them. We provide a systematic and critical review of hybrid fairness approaches, organizing existing methods according to the fairness mechanisms they employ and the algorithmic and mathematical strategies used to reconcile multiple fairness criteria. For each class of methods, we examine their theoretical foundations, optimization mechanisms, and empirical evaluation practices, and discuss their limitations. Additionally, we discuss the challenges and identify open research directions for developing principled, context-aware hybrid fairness methods. 
By synthesizing insights across the literature, this survey aims to serve as a comprehensive resource for researchers and practitioners seeking to design hybrid algorithms that provide reliable fairness guarantees at both the individual and group levels.
\end{abstract}

\section{Introduction}
\label{sec:intro}

The increasing reliance on Machine Learning (ML) for high-stakes decision-making across domains, including criminal justice, loan approvals, hiring, and healthcare, has heightened concerns about algorithmic fairness \cite{chiang2022exploring}. While Artificial Intelligence (AI) offers significant efficiency gains, it risks systematizing and scaling historical biases, leading to measurable real-world harm \cite{barocas2023fairness}.  These concerns are central to Responsible AI, where the deployment of automated decision systems must account not only for performance, but also for normative, legal, and societal implications \cite{papagiannidis2025responsible}. A guiding principle in this domain is the search for the ``less discriminatory alternative'', meaning that among equally effective algorithms, the one imposing the least disparate impact should be preferred. As Laufer et al. \cite{laufer2025constitutes} argue, identifying such alternatives requires methods that rigorously combine both quantitative metrics and qualitative perspectives.
This plurality has given rise to a vast literature on algorithmic fairness, encompassing bias identification, measurement, and mitigation. 

Within this literature, fairness notions are often divided into two conceptually distinct families: \emph{Group Fairness} (GF) and \emph{Individual Fairness} (IF). 
GF focuses on parity of outcomes or error rates across groups defined by sensitive attributes, while IF emphasizes consistency of treatment among similar individuals, regardless of group membership. Despite their shared ethical motivation, these two perspectives impose fundamentally different and often conflicting constraints on learning systems. As emphasized in surveys such as \cite{mehrabi2021survey}, definitions of fairness vary widely across disciplines and societal contexts.  However, comparatively little attention has been paid to how conflicts between fairness notions shape both technical design choices and downstream societal consequences when multiple criteria are considered simultaneously.

Most existing work studies GF and IF largely in isolation. 
Research is typically organized either around specific fairness metrics or around the stage of the ML pipeline at which fairness interventions occur. A common taxonomy distinguishes stages of intervention: \emph{pre-processing}, in which input data is modified to meet fairness criteria; \emph{in-processing}, in which the model architecture is modified during training to ensure fairness; and \emph{post-processing}, in which model predictions are adjusted. This taxonomy is widely used because it links methodological decisions to practical constraints, such as data availability, retraining costs, and regulatory limitations. For a broad overview of this landscape, Hort et al. \cite{hort2024bias} provide a comprehensive review analyzing 341 studies on bias mitigation techniques through this framework. However, this methodological perspective does not fully capture how different \emph{definitions} of fairness shape both the design and the consequences of such interventions. Specifically, this pipeline-centered perspective largely abstracts away from how different definitions of fairness condition the nature of resulting trade-offs.

Beyond where fairness interventions are applied in the pipeline, a more fundamental distinction concerns \emph{how fairness itself is defined}. IF relies on the principle that ``similar individuals are treated similarly'' \cite{dwork2012fairness}. Implementing this idea typically requires defining a distance or similarity function between individuals, which can be challenging and highly domain-dependent. GF, on the other hand, is generally computed by comparing statistical quantities across predefined groups. These metrics can be directly estimated from model outputs and observed outcomes, making GF relatively straightforward to compute once sensitive attributes are available.  

As a result, IF has historically been less common in empirical and applied work because of the difficulty of specifying appropriate similarity metrics, despite its conceptual alignment with the ethical principle of equitability. While recent advances have explored learned, causal, or representation-based notions of individual similarity, IF remains underrepresented in real-world deployments compared to group-based approaches, as noted in reviews such as \cite{bernard2025systematic}.

\begin{figure}[ht!]
    \centering
    \includegraphics[width=\linewidth]{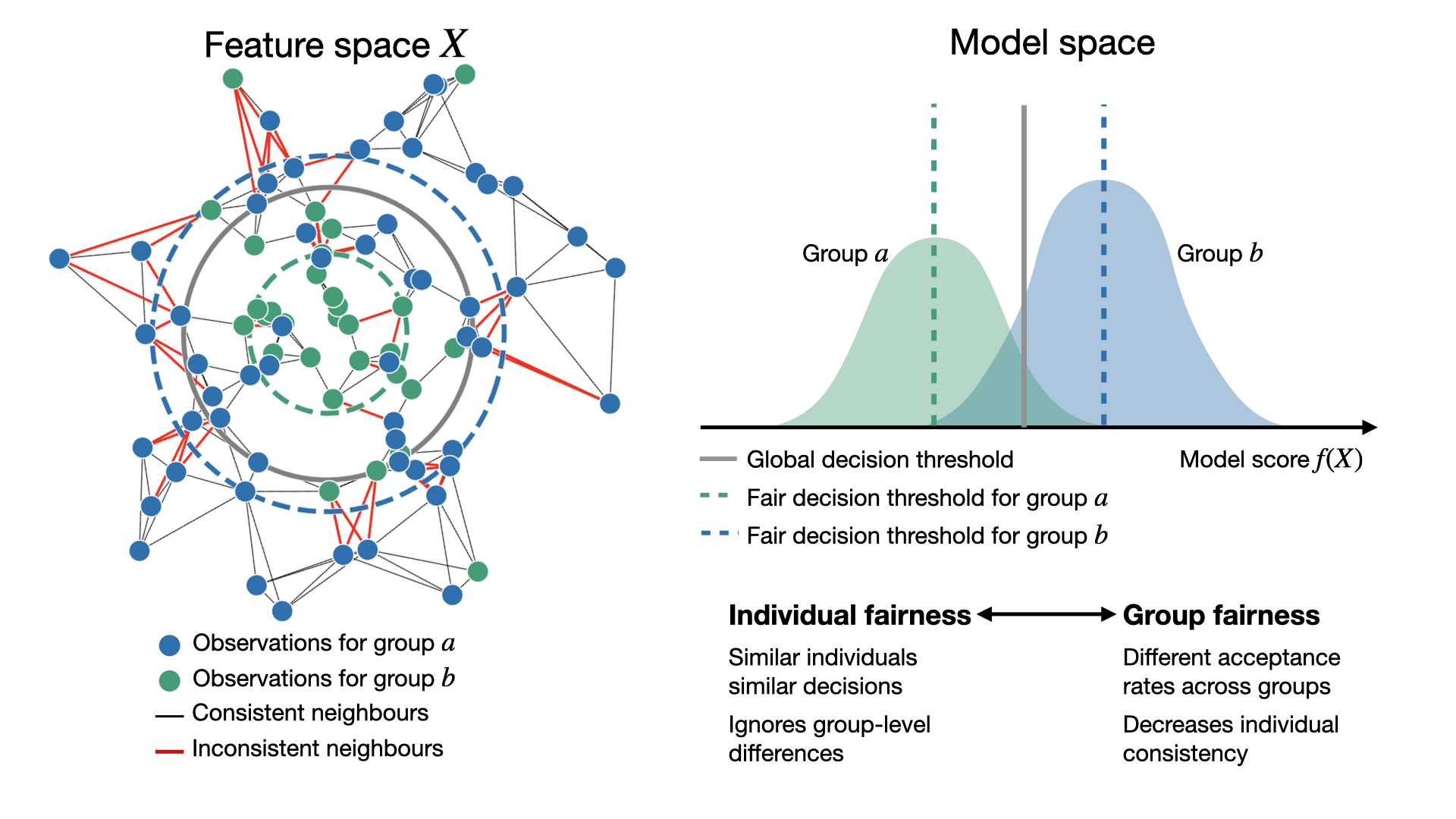}
    \caption{\textbf{Trade-off between group and individual fairness.} \textit{Left}: observations from two groups in feature space $X$, where edges indicate similar individuals (using $k$-nearest neighbors) who should receive similar decisions (individual fairness). A single global threshold (gray radius) maximizes local consistency but yields different acceptance rates across groups. Group-specific thresholds (dashed radii) enforce Independence (equal acceptance rates) but break consistency across similar neighbors (red edges). \textit{Right}: the same thresholds shown in model-score space $f(X)$, where differing score distributions make these fairness criteria incompatible.}
    \label{fig:trade-offs_diagram}
\end{figure}

Critically, optimizing for one type of fairness often compromises the other. Although GF and IF share the same objective of equitable treatment, they are often mathematically incompatible. Achieving perfect group parity may require treating individuals differently, thus violating IF, while enforcing strict IF can preserve group-level disparities \cite{dwork2012fairness, anderson2025algorithmic}. Recent theoretical studies have formalized these trade-offs: \cite{xu2024compatibility} shows that under common conditions, GF and IF cannot be simultaneously satisfied except in trivial cases, and \cite{zhou2022group} provides empirical evidence that improving GF can worsen IF in practice. These incompatibilities are not merely technical artifacts, but reflect deeper normative tensions about which dimensions of fairness should be prioritized in socio-technical systems. Figure \ref{fig:trade-offs_diagram} illustrates the core trade-off between GF and IF: enforcing group-level parity introduces local inconsistencies among similar individuals, while enforcing individual consistency can reproduce group-level disparities.

Unlike existing surveys that focus broadly on bias mitigation or prioritize a single definition (e.g. \cite{caton2024survey}), we focus explicitly on methods that leverage both GF and IF and make the resulting trade-offs explicit. Thus, this paper provides a comprehensive review of approaches that integrate both notions within a unified framework, highlighting how different design choices balance competing fairness objectives. To the best of our knowledge, existing surveys do not systematically analyze methods that explicitly manage or expose GF–IF trade-offs across definitions, algorithms, and evaluation settings.

The remainder of this work is organized as follows: Section~\ref{sec:preliminaries} reviews key literature on Fairness in Machine Learning, focusing on existing surveys in the field. It also establishes the core terminology, notation, and definitions used throughout the paper. Section~\ref{sec:trade-off} reviews works that address GF and IF simultaneously, explicitly managing the trade-offs between these two concepts. This Section also provides summary tables of the surveyed algorithms. Finally, Section~\ref{sec:discusion} highlights the main conclusions and limitations, and identifies directions for future work.

\section{Preliminaries}
\label{sec:preliminaries}

This section establishes the core terminology and notation necessary for analyzing the trade-offs between Group Fairness (GF) and Individual Fairness (IF), which were merely noted in Section~\ref{sec:intro}. 

We first review the key literature surrounding fairness surveys to contextualize the state of the art. We then introduce the unified notation used throughout the manuscript. Establishing a consistent mathematical language is essential for comparing methods originating from different communities and embodying distinct modeling assumptions. Next, we formally define GF and IF in the context of machine learning. These definitions constitute the conceptual foundation for much of the existing literature. Finally, we describe the principles that characterize the main families of fairness intervention methods (pre-processing, in-processing, and post-processing). 
These preliminaries lay the groundwork for understanding both the theoretical and practical trade-offs discussed in subsequent sections.

\subsection{Related Surveys}
\label{sec:related_surveys}

Recent literature has produced several systematic reviews and surveys that categorize fairness metrics, mitigation strategies, and existing tensions. Key contributions include:

\begin{itemize} 

\item Bernard and Balog (2025) \cite{bernard2025systematic}: A systematic review of fairness, accountability, transparency, and ethics in information retrieval. It specifically notes that while GF and IF are mentioned, the latter remains largely unexplored in this domain. It suggests that investigating the trade-offs between IF and GF is a crucial topic for future research. 

\item Gao et al. (2025) \cite{gao2025fair}: A systematic review of fairness in machine learning (ML) with a focus on health applications. It examines why ML models are unfair and how to measure fairness in real-world applications. It reviews fairness contributions within GF, IF, and causal-based fairness. 

\item Rabonato and Berton (2025) \cite{rabonato2025systematic}: A systematic review of fairness in ML using ACM, IEEE, and Springer databases. It provides a broad overview of definitions, metrics, and mitigation strategies, explicitly highlighting conflicts among fairness definitions and the fairness-accuracy trade-off. 

\item Caton and Haas (2024) \cite{caton2024survey}: A comprehensive survey of fairness in ML covering notions of fairness, types of bias, and practical challenges. It organizes mitigation approaches into pre-processing, in-processing, and post-processing methods, and discusses the inherent conflict between fairness and accuracy. 

\item Hort et al. (2024) \cite{hort2024bias}: A systematic review focusing on bias mitigation methods. It categorizes a vast number of publications into pre-processing (123), in-processing (212), and post-processing (56), while also addressing the fairness-accuracy trade-off. 

\item Pagano et al. (2023) \cite{pagano2023bias}: A systematic review using Scopus, IEEE Xplore, Web of Science, and Google Scholar. It provides an overview of fairness definitions and metrics, noting the frequent conflicts between different fairness definitions and predictive performance. 

\item Alves et al. (2023) \cite{alves2023survey}: A survey focused on formalizing fairness notions and discussing related tensions. It analyzes conflicts among different fairness metrics and their tensions with other desirable properties, such as privacy and accuracy. 

\item Mehrabi et al. (2021) \cite{mehrabi2021survey}: A fundamental survey listing sources of biases that affect AI systems. It focuses on the origins of bias, fairness metrics, and mitigation strategies, categorizing interventions into pre-processing, in-processing, and post-processing methods. \end{itemize}

\subsection{Notation}
\label{sec:notation}

Let $\mathcal{D} = \{(\mathbf{x}_i, \mathbf{s}_i, y_i)\}_{i=1}^N$ denote a dataset of $N$ observations, where $\mathbf{s}_i \in S$ represents the sensitive attributes of observation $i$,  $\mathbf{x}_i \in X$ its non-sensitive features, and  $y_i \in Y$ the target variable. A statistical model $f(\mathbf{x}_i, \mathbf{s}_i)$ produces a prediction $\hat{y}_i$,  which can be interpreted either as a real-valued output in regression or as a score in classification.  In classification settings, the predicted class is given by the label with the highest score, $\hat{y}_i = \arg \max\limits_{c\in\mathcal{C}} f_c(\mathbf{x}_i, \mathbf{s}_i)$, where $\mathcal{C}$ is the set of classes and $f_c(\mathbf{x}_i, \mathbf{s}_i)$ denotes the score assigned to class $c$. For simplicity, we focus on \emph{binary classification with a single binary sensitive attribute}, i.e., $Y=\{0,1\}$ and $S=\{0,1\}$,
 where a single threshold $\tau$ maps the score to a class label, $\hat{y}_i = \mathbb{I}\!\left[f(\mathbf{x}_i,\mathbf{s}_i) \ge \tau\right]$, with $\mathbb{I}\left[\cdot\right]$ denoting the indicator function. This setting allows fairness notions to be expressed in the most transparent way.  Most fairness definitions and results extend naturally to multi-class classification, regression, and multiple sensitive attributes; see \cite{zhou2022group}.
\
\subsection{Types of fairness}
\label{sec:types_of_fairness}

As mentioned earlier, fairness in machine learning is typically addressed through two distinct conceptual lenses: GF, which aims for collective outcomes for protected populations, and IF, which aims for consistent treatment of specific instances.

As highlighted by \cite{anderson2025algorithmic}, these two frameworks treat justice differently: GF focuses on equalizing error rates or benefits across groups, while IF enforces that similar individuals receive similar predictions. Importantly, these notions are not merely different but often mathematically incompatible. Satisfying strict group parity often requires treating similar individuals differently based on their group membership to correct for distributional imbalances, thereby violating individual fairness constraints \cite{zhou2022group}. 

\subsubsection{Group Fairness} 
\label{sec:def_group_fairness}

Group fairness requires that an algorithm achieves comparable performance across sensitive groups \cite{anderson2025algorithmic}. Intuitively, a GF algorithm should not systematically favor or disadvantage individuals based on their group membership. GF is therefore evaluated using \emph{group-level metrics}, and the choice of metric determines what “fairness” means in practice.

There are three main criteria commonly used to evaluate GF in algorithmic decision-making \cite{kozodoi2022fairness}. The first, and most widely adopted, is \emph{independence}, which requires predictions to be statistically independent of sensitive attributes,
\begin{equation}
    \hat{Y} \perp S.
    \label{eq:gf_independence}
\end{equation}
A standard way to operationalize independence is through \emph{demographic parity}, also known as statistical parity, introduced by \cite{calders2010three}, which requires
\begin{equation}
    P(\hat{Y} = 1 \mid S = 0) - P(\hat{Y} = 1 \mid S = 1) \approx 0,
    \label{eq:demographic_parity}
\end{equation}
where $\hat{Y} = 1$ denotes a positive prediction and $S=1$ indicates membership in the sensitive group.

The second criterion is \emph{separation}, which also requires statistical independence from the sensitive attributes, but conditioned on the true outcome,
\begin{equation}
    \hat{Y} \perp S \mid Y.
    \label{eq:gf_separation}
\end{equation}
Separation is commonly enforced through \emph{equalized odds} \cite{hardt2016equality}, which requires both true positive and false positive rates to match across groups,
\begin{equation}
    \frac{1}{2} \sum_{y \in \{0,1\}} \Big[ P(\hat{Y}=1 \mid Y=y, S=0) - P(\hat{Y}=1 \mid Y=y, S=1) \Big] \approx 0.
    \label{eq:equalized_odds}
\end{equation}
A relaxed version of this criterion is \emph{equal opportunity}, which requires equality only of true positive rates across groups.

The final criterion is \emph{sufficiency}, which requires the target variable to be statistically independent of sensitive attributes given the prediction,
\begin{equation}
    Y \perp S \mid \hat{Y}.
    \label{eq:gf_sufficiency}
\end{equation}
Compared to separation, the direction of conditioning is reversed. Sufficiency can be operationalized through \emph{predictive parity}, which requires the positive predictive value to match across sensitive groups,
\begin{equation}
    P(\hat{Y} = 1 \mid Y=1, S = 0) - P(\hat{Y} = 1 \mid Y=1, S = 1) \approx 0,
    \label{eq:predictive_parity}
\end{equation}
Unlike demographic parity, criteria based on separation and sufficiency explicitly account for predictive performance relative to the ground truth. 

Many other group-level criteria have been proposed (e.g., reputation disparity \cite{ramos2022robust} and exposure-based measures \cite{zehlike2020reducing}), but demographic parity, equalized odds, equal opportunity, and predictive parity remain the most widely used and representative notions of GF. These criteria capture the core principle of GF: enforcing comparable model behavior across groups according to a predefined fairness objective.

\subsubsection{Individual Fairness} 
\label{sec:def_individual_fairness}

One of the most widely adopted definitions of Individual Fairness states that similar individuals should receive similar predictions  \cite{dwork2012fairness}. In statistical terms, similarity is defined using non-sensitive features. Thus, if two individuals are close in feature space, their predicted outcomes should also be close.  

A general operational definition of IF \cite{friedler2021possibility} uses distance metrics. An algorithm is $(\epsilon, \epsilon')$-fair if, for any two individuals $i$ and $j$,
\begin{equation}
    d_F(\bm{x}_i, \bm{x}_j) \leq \epsilon \ \implies \ d_P(\hat{y}_i, \hat{y}_j) \leq \epsilon',
    \label{eq:individual_fair_ee}
\end{equation}
where $d_F$ and $d_P$ are distance metrics in the feature and prediction spaces, respectively. In other words, individuals who are close in their non-sensitive attributes must receive predictions that are correspondingly close. Equivalently, this condition can be expressed as a Lipschitz continuity requirement,
\begin{equation}
    d_P(\hat{y}_i, \hat{y}_j)\leq L\, d_F(\bm{x}_i,\bm{x}_j),
    \label{eq:individual_fair_lipschitz}
\end{equation}
for some Lipschitz constant $L>0$, ensuring that small differences in relevant attributes induce bounded differences in predictions. Some authors further refine this notion by adopting a counterfactual interpretation of individual fairness (see, e.g., \cite{han2023dualfair, lohia2019bias}).  Let $\bm{x}_i=(\bm{x}_i, s_i)$ denote the feature vector of individual $i$. The counterfactual counterpart of individual $i$ is defined as
\[
\bm{x}_i^{\mathrm{cf}} = (\bm{x}_i, 1-s_i),
\]
that is, a hypothetical version of the same individual that is identical in all non-sensitive characteristics but differs only in the value of the sensitive attribute. Under this perspective, individual fairness requires prediction invariance, or at least bounded variation, across such counterfactual pairs, namely
\begin{equation}
    d_P\big(f(\bm{x}_i), f(\bm{x}_i^{\mathrm{cf}})\big) \leq \epsilon'.
\end{equation}
This formulation can be seen as a special case of the Lipschitz-based definition.

Often, one can learn a transformation $\tilde{\bm{x}}_i = \phi(\bm{x}_i, \bm{s}_i)$ that maps both sensitive and non-sensitive features into a new representation. The transformation $\phi$ satisfies individual fairness if the distances between transformed and original features remain close, i.e., $\big| d_F(\tilde{\bm{x}}_i, \tilde{\bm{x}}_j) - d_F(\bm{x}_i, \bm{x}_j) \big| \leq \epsilon.$
In this case, individuals who are similar in their non-sensitive attributes remain similar in the transformed representation \cite{lahoti2019ifair}.

\subsection{Stages of processing}
\label{sec:stages_of_processing}

The strategies for achieving fairness and mitigating bias in machine learning are often classified into three categories: \emph{pre-processing}, \emph{in-processing}, and \emph{post-processing} methods \cite{pessach2022review, hort2024bias, ferrara2024fairness, shah2025comprehensive}. These categories differ in the timing of the fairness intervention within the machine learning pipeline.

\subsubsection{Pre-processing} 
Pre-processing (PRE) methods modify the training data \textit{before} fitting the model. The goal is to reduce correlations between sensitive attributes and the target variable or features, thereby preventing bias from propagating into the model. Common approaches include data reweighting, resampling, or representation learning that obfuscates sensitive information while preserving predictive signal. For example, rebalancing datasets can ensure comparable representation across sensitive groups, while learned transformations can produce fair feature embeddings \cite{pessach2022review, hort2024bias}.

\subsubsection{In-processing} 
In-processing (IN) methods often modify the model architecture by incorporating fairness constraints or objectives directly into the learning algorithm \textit{during} training. This typically involves adding fairness regularizers to the loss function, modifying optimization procedures, or using adversarial debiasing techniques. These methods look to balance predictive accuracy with fairness guarantees in a principled way. In-processing approaches are considered powerful because they integrate fairness into the core learning dynamics. However, they can also be computationally demanding, model-specific, and more challenging to integrate into the machine learning pipeline, as it requires modifying the model itself \cite{hort2024bias, shah2025comprehensive}.

\subsubsection{Post-processing} 
Post-processing (POST) methods adjust model outputs after training, without modifying the data or the model itself. These approaches learn a mapping from predictions to adjusted outputs that satisfy fairness criteria. Examples include group-specific decision thresholds, calibrated equalized-odds adjustments, or rejection-option classification. Post-processing methods are attractive because they treat the model as a black box and can be applied to any trained predictor, but they often involve trade-offs in interpretability or predictive power \cite{ferrara2024fairness, hort2024bias}.

\section{Trade-offs between group and individual fairness}\label{sec:trade-off}


Ensuring fairness in algorithmic decision-making requires addressing both group-level and individual-level considerations, as these notions capture \textit{complementary} and \textit{non-redundant} aspects of justice. Group fairness (GF) aims to prevent discrimination at the population level, but satisfying group-level criteria alone does not guarantee fair treatment of individuals. Conversely, individual fairness (IF) enforces consistency by requiring that similar individuals be treated similarly; however, when applied in isolation, it may inadvertently perpetuate existing social inequalities if the underlying feature space or similarity metric reflects historical or societal biases. As a result, neither notion is sufficient on its own, and their joint consideration is essential for a more comprehensive and ethically defensible understanding of algorithmic justice.

 Importantly, prior work has shown that GF and IF are not only conceptually distinct but can also be formally incompatible under standard definitions \cite{dwork2012fairness, xu2024compatibility}. This tension can be illustrated through a simple post-processing argument. Suppose IF is enforced via the Lipschitz condition in Eq.~(\ref{eq:individual_fair_lipschitz}), while GF is defined through independence, as in Eq.~(\ref{eq:gf_independence}). Assume further that fairness is imposed through post-processing, keeping the original model fixed and adjusting predictions to satisfy group-level constraints. In this setting, the optimal GF intervention corresponds to a transformation $T$ that minimally alters the original predictions while enforcing independence:
 
\begin{equation*}
    T = \arg\min_{\tilde{T}} \left\{ \lVert \hat{Y} - \tilde{T}(\hat{Y}, S) \rVert^2 : \tilde{T}(\hat{Y}, S) \perp S \right\}.
\end{equation*}
If the original predictions are not group-fair (i.e., $\hat{Y} \not\perp S$), achieving independence necessarily requires the transformation to depend on the sensitive attribute. As a result, there exist prediction values $\hat{y}$ and sensitive attributes $s_1 \neq s_2$ such that
\begin{equation*}
    T(\hat{y}, s_1) \neq T(\hat{y}, s_2).    
\end{equation*}
This immediately violates IF: two individuals with identical features and identical original predictions receive different outcomes solely due to their group membership. Formally, the Lipschitz condition fails,
\begin{equation*}
    d_P\big(T(\hat{y}, s_1), T(\hat{y}, s_2)\big) > 0 = L\, d_F(\hat{y}, \hat{y}),
\end{equation*}
for any Lipschitz constant $L > 0$.

This reasoning shows that enforcing GF through independence necessarily introduces individual-level inconsistencies whenever group disparities are present. More generally, this tension is not limited to post-processing or exact statistical parity: analogous incompatibilities arise across learning stages and persist even when fairness constraints are relaxed, manifesting as trade-offs along the Pareto frontier between GF and IF objectives \cite{joseph2016fairness, corbett2023measure, xu2024compatibility}.

\begin{figure}[ht!]
    \centering
    \begin{subfigure}[t]{0.48\linewidth}
        \centering
        \includegraphics[width=\linewidth]{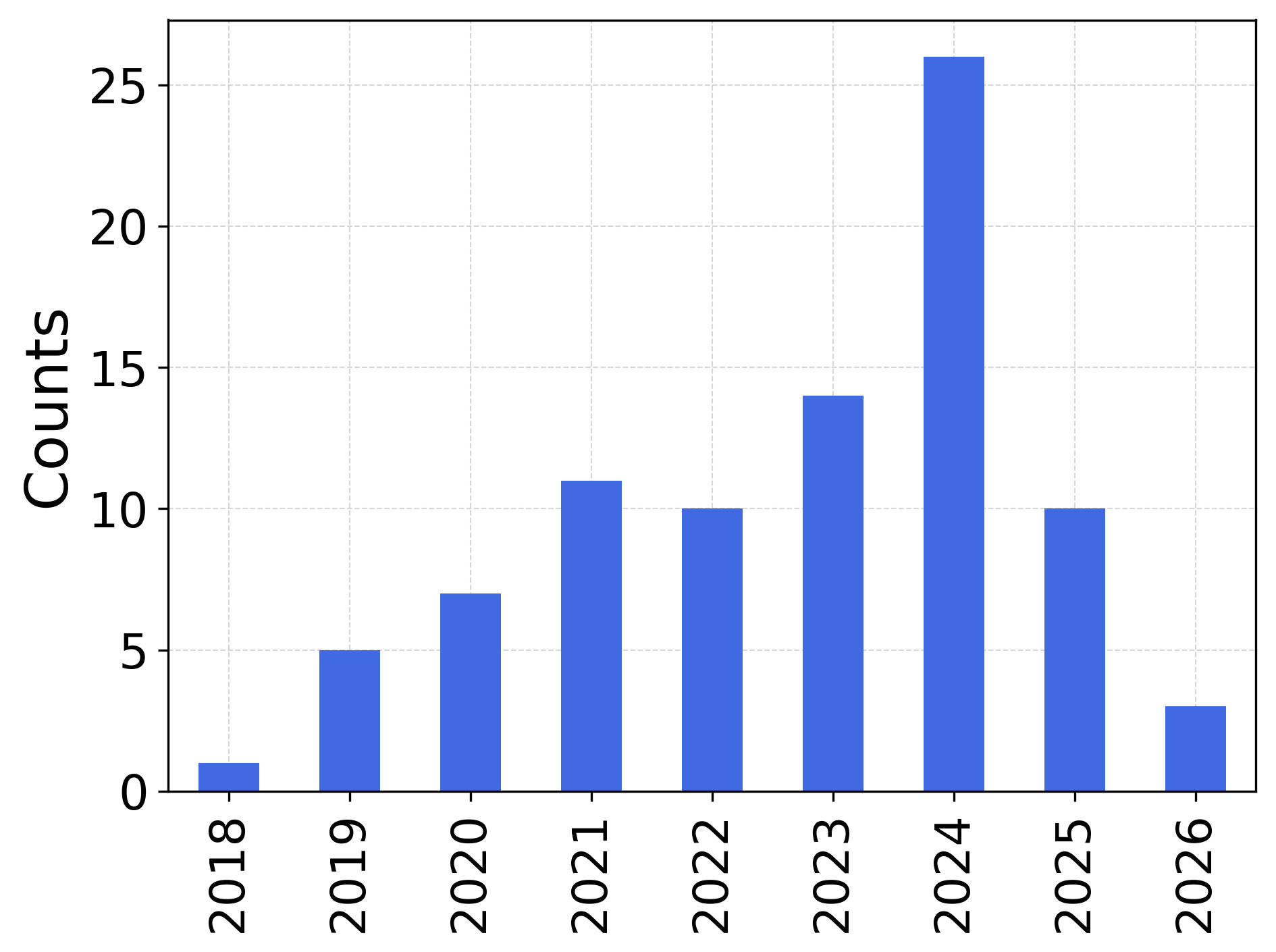}
        \caption{Excluding arXiv.}
        \label{fig:papers_year_woarxiv}
    \end{subfigure}
    \hfill
    \begin{subfigure}[t]{0.48\linewidth}
        \centering
        \includegraphics[width=\linewidth]{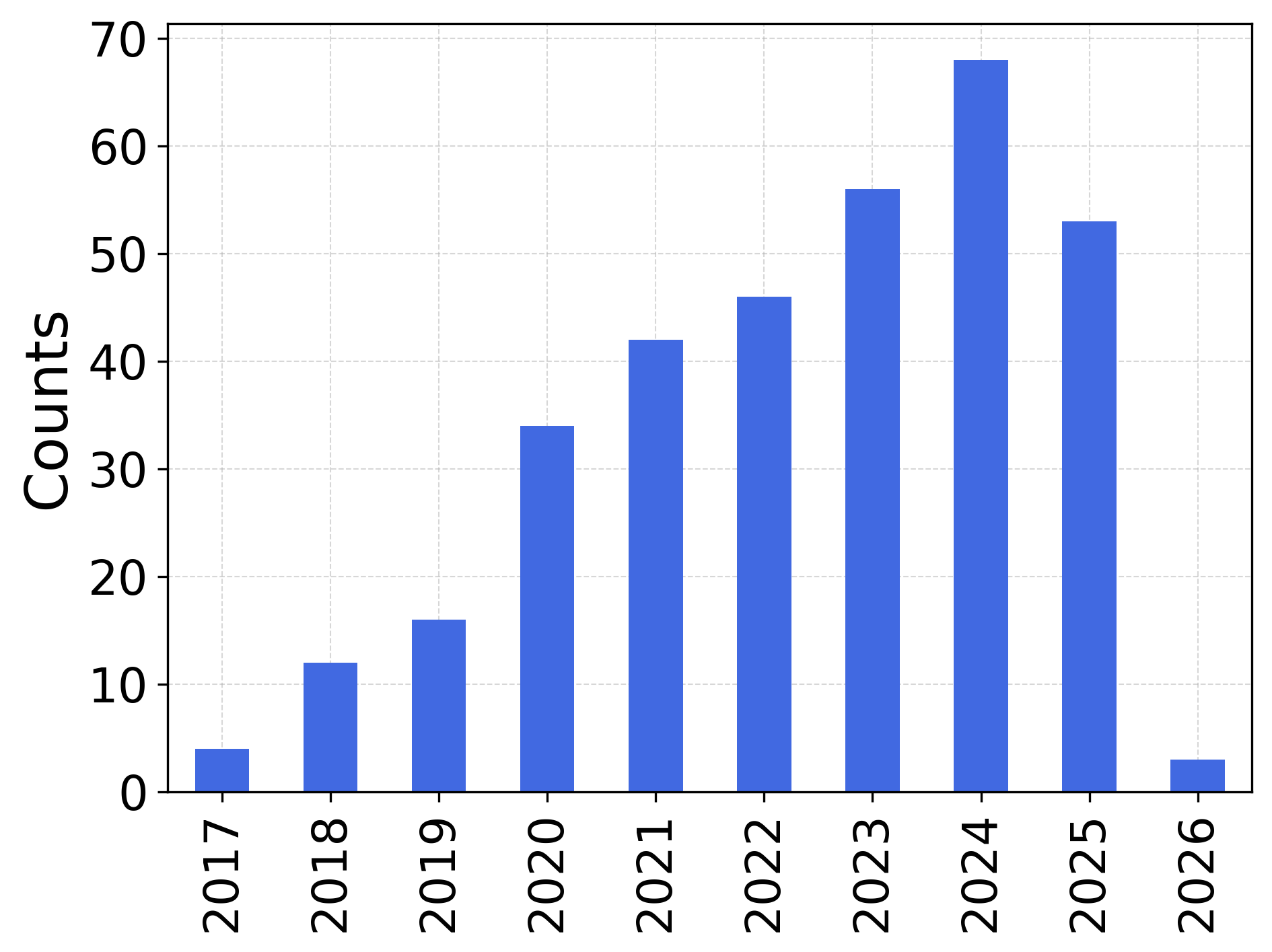}
        \caption{Including arXiv.}
        \label{fig:papers_year_arxiv}
    \end{subfigure}

    \caption{Number of papers identified per year through the literature search process. Panel (a) shows results excluding arXiv, while panel (b) includes arXiv preprints.}
    \label{fig:papers_year_comparison}
\end{figure}

Motivated by these arguments, this section provides a comprehensive analysis of methodologies that aim to address GF and IF simultaneously, outlining how recent approaches attempt to balance these two foundational yet often competing principles. To this end, a systematic literature search was conducted to identify studies that address both IF and GF in machine learning. The search was performed across different bibliographic sources, including the Web of Science\footnote{\url{https://www.webofscience.com/wos/alldb/advanced-search}}, the IEEE repository\footnote{\url{https://ieeexplore.ieee.org/search/advanced}}, or arXiv\footnote{\url{https://arxiv.org/search/advanced}}, using a topic-based Boolean query combining terms related to IF and GF:
{\footnotesize
\begin{verbatim}
("individual fairness" OR "individual-level fairness" OR consistency OR "lipschitz fairness")
AND
("group fairness" OR "demographic parity" OR "statistical parity"
 OR "equalized odds" OR "equal opportunity")
\end{verbatim}
}
Figures~\ref{fig:papers_year_woarxiv} and \ref{fig:papers_year_arxiv} display the number of identified papers per year, excluding and including arXiv publications, respectively. ArXiv papers are excluded in Figure~\ref{fig:papers_year_woarxiv} to focus on peer-reviewed and formally published research, while Figure~\ref{fig:papers_year_arxiv} incorporates preprints to capture the most recent and emerging contributions that have not yet completed the publication process. In both cases, a clear upward trend is observed, indicating a growing and sustained research interest in this topic over time.

The objective of this search was to capture studies that discuss both notions within the same contribution. During the screening process, it was observed that a substantial portion of the retrieved papers mention both notions in the introduction or related work sections as part of a general motivation or positioning within the fairness literature, without formalizing a joint optimization, constraint, or trade-off between the two fairness paradigms in the proposed methodology. Consequently, an additional filtering step was applied to retain exclusively those papers that explicitly integrate both notions into the learning pipeline by formalizing a trade-off, joint objective, constraint set, or evaluation framework that simultaneously accounts for group and individual fairness.

The final set of selected papers is summarized in Table~\ref{tab:summ_1}, while a more detailed discussion of the fairness mechanisms and methodological approaches adopted by each work is provided in Section~\ref{subsec:taxonomy}. In addition, our findings are organized into two complementary tables that review the same core body of literature from different perspectives. Specifically, Table~\ref{tab:table_trade-offs} presents a methodological taxonomy, classifying approaches according to their intervention stage in the learning pipeline and the specific sub-techniques employed. Complementarily, Table~\ref{tab:table_datasets} builds directly on this taxonomy by shifting the focus to the practical implementation of these methods, summarizing reported computational costs and the benchmark datasets used for empirical validation. Taken together, these tables offer both a structured theoretical overview and a practical assessment of the current state of the literature.

\begin{table}[ht!]
    \centering
    \begin{tabular}{|l|l|}
    \hline
        \textbf{Bias Control Mechanism} & \textbf{References} \\
        \hline
         Data transformation &  \cite{calmon2017optimized, xu2024compatibility} \\
         Reweighting and resampling &  \cite{arnaiz2023towards} \\
         Representation and embedding &  \cite{zemel2013, liu2025leveraging, han2023dualfair} \\
         Augmentation and Contrastive & \cite{calmon2017optimized, han2023dualfair, li2022accurate,lohia2019bias}\\
         Regularization &  \cite{liu2025leveraging, han2023dualfair, li2022accurate, nagpal2025optimizing, wang2024individual2, wang2024individual, liu2024adversarial, long2023individual, dwork2012fairness} \\
         Metric-based approaches &  \cite{li2022accurate, dwork2012fairness,small2024equalised} \\
         Adversarial learning &  \cite{liu2024adversarial} \\
         Thresholding & \cite{ small2024equalised}\\
         Reject-option & \cite{lohia2019bias}\\
         \hline
    \end{tabular}
    \caption{Summary of references by type of applied method}
    \label{tab:summ_1}
\end{table}

\subsection{Taxonomy}
\label{subsec:taxonomy}

A wide variety of approaches has been proposed to simultaneously address GF and IF within machine learning pipelines. These approaches are often classified by the stage at which the fairness intervention is applied (pre-processing, in-processing, or post-processing), as reviewed in Section~\ref{sec:stages_of_processing}. While this classification is useful from a pipeline-oriented perspective, it does not fully capture that many fairness strategies are inherently stage-agnostic and can be instantiated at different points in the learning process. For instance, mechanisms such as data reweighting or data augmentation may appear as pure pre-processing (PRE) techniques or be embedded directly into the learning objective (IN).

For this reason, beyond the standard stage-based categorization (i.e., following the notation used by \cite{caton2024survey}), we adopt a fine-grained taxonomy that organizes methods by the type of fairness mechanism they introduce, rather than by the specific pipeline stage at which they are deployed. The proposed taxonomy groups methods into representative families based on how they formulate and manage the GF-IF trade-off, how fairness objectives or constraints are defined, and how fairness-aware components are incorporated into the data representation, the optimization process, or the model outputs. This perspective highlights common conceptual principles shared by methods that may otherwise be assigned to different stages of the learning pipeline. 

  \paragraph{\textbf{Strategies based on representation or data modification.}}
    \begin{itemize}
     
      \item \emph{Data transformation}: These methods operate directly on the input data $(X,S,Y)$ to mitigate bias before model training. They include procedures such as \textit{label correction} or \textit{feature modification}, designed to balance the joint distribution of sensitive groups and outcome labels. Specifically, they aim to produce a transformed dataset $\tilde{\mathcal{D}} = (\tilde{x}_i,s_i,\tilde{y}_i)_{i=1}^N$ such that the resulting empirical distribution $\tilde{P}(\tilde{X},S,\tilde{Y})$ exhibits reduced dependence between $S$ and $\tilde{Y}$ (or $S$ and $\tilde{X}$) while maintaining fidelity to the original data-generating process. 
      From a statistical standpoint, these procedures can be interpreted as approximating samples from a counterfactual distribution that satisfy fairness constraints (such as demographic parity or equalized odds). Techniques in this category include data relabeling via conditional probability calibration or feature editing guided by causal or disentanglement criteria.
      Data transformation primarily fosters \textit{group fairness} by equalizing representation or outcomes across sensitive groups. However, they may also enhance \textit{individual fairness} when the transformation preserves neighborhood relations in the feature space. A representative method of this paradigm is \cite{calmon2017optimized}, which illustrates how data transformations can be designed to equalize outcomes across sensitive groups while preserving local neighborhood structures, thereby supporting both group and individual fairness. Another representative method in this line of work is \cite{xu2024compatibility}.

       \item \emph{Reweighting and resampling}: These approaches modify the observed sample distribution $P(X, S, Y)$ rather than altering the data directly. They assign instance-specific weights $w_i$ such that the resulting \textit{weighted empirical distribution}
\[
\tilde{P}_w(X, S, Y) = \sum_{i=1}^N w_i\, \delta_{(X_i, S_i, Y_i)}
\]
approximates a target distribution in which $S$ and $Y$ (or $S$ and $f(X)$) are approximately independent. 
Here, $\delta_{(X_i, S_i, Y_i)}$ denotes a \textit{Dirac delta measure} centered at the observation $(X_i, S_i, Y_i)$, i.e., a point mass assigning all probability to that specific instance.

From a statistical perspective, this can be viewed as \textit{importance weighting under fairness constraints}. The corresponding \textit{weighted empirical risk},
\[
R_w(f) = \sum_i w_i\, L(f(X_i), Y_i),
\]
serves as an estimator of a target risk under a ``fair'' distribution, where $P(S)$ is balanced or the conditional dependence between $S$ and $Y$ is attenuated. 

Different formulations impose these fairness constraints in various ways. Some methods explicitly minimize covariance terms (e.g., enforcing $\operatorname{Cov}(S, f(X)) \approx 0$), while others derive $w_i$ using Shapley values or influence-function analyses that quantify each instance’s marginal contribution to overall group-level disparity.

Reweighting primarily promotes \textit{group fairness} by adjusting the relative importance of instances from different sensitive subpopulations. At the same time, certain reweighting schemes can also influence \textit{individual fairness}, for example, by modifying how similar individuals are weighted or treated during training. A seminal approach is \cite{arnaiz2023towards}, which focuses on mitigating global, group-level disparities through sample reweighting. Here, the authors use Shapley values to assess the contribution of individual observations to the fairness criteria, providing an individual-level perspective without explicitly enforcing local notions of fairness.

      \item \emph{Representation and embedding}: 
These approaches modify the data representation by mapping individuals into a latent space in which similar observations (according to task-relevant and fairness-aware criteria) are grouped and represented by shared prototypes or embeddings. Rather than reweighting instances or altering raw feature values, representation-based methods learn a transformation
\[
\phi:\mathcal{X}\rightarrow\mathcal{Z},
\]
such that individuals who are close in the original feature space and deemed similar under a fairness-aware metric are mapped to nearby points in the representation space $\mathcal{Z}$.

A common strategy involves selecting or learning \emph{representative instances} (e.g., prototypes, centroids, or barycenters) for clusters of similar observations, and training the predictor using these representatives instead of the original samples. Formally, given a partition $\{\mathcal{C}_k\}_{k=1}^K$ of the dataset, each cluster $\mathcal{C}_k$ is summarized by a representative $\tilde{x}_k$, and the empirical distribution is approximated by
\[
\tilde{P}_{\mathrm{rep}}(X,S,Y) = \sum_{k=1}^K \pi_k\, \delta_{(\tilde{x}_k,\tilde{s}_k,\tilde{y}_k)},
\]
where $\pi_k$ denotes the relative mass of cluster $\mathcal{C}_k$ and $(\tilde{s}_k,\tilde{y}_k)$ are aggregated or inferred labels.

From a fairness perspective, representation learning provides a natural mechanism to enforce IF, as individuals within the same cluster are explicitly treated as indistinguishable by the model, leading to similar predictions by construction. At the same time, GF can be promoted by imposing constraints on the representation itself, such as limiting the recoverability of sensitive attributes from $\phi(X)$ or enforcing balanced cluster compositions across sensitive groups.

Representative approaches, including \cite{zemel2013, liu2025leveraging, han2023dualfair}, illustrate how learning fair representations can simultaneously control group-level disparities while preserving individual-level consistency through representative-based abstraction.

      \item \emph{Augmentation and Contrastive}: 
These approaches expand the observed dataset by constructing synthetic but structured instances that represent alternative realizations of the same individual under controlled perturbations, typically involving the sensitive attributes. Rather than modifying existing observations or their weights, these methods augment the empirical distribution by introducing contrastive or counterfactual samples designed to enforce fairness constraints.

Formally, let $(X_i, S_i, Y_i)$ denote an observed instance, where $X_i$ are non-sensitive features and $S_i$ is a sensitive attribute. A counterfactual or contrastive counterpart is defined as
\[
(X_i, S_i', Y_i^{\mathrm{cf}}), \quad S_i' \neq S_i,
\]
where $Y_i^{\mathrm{cf}}$ may be inherited, imputed, or left unlabeled depending on the method. The augmented empirical distribution can then be written as
\[
\tilde{P}_{\mathrm{aug}}(X,S,Y) = \frac{1}{N + N_{\mathrm{cf}}}
\sum_{i=1}^{N+N_{\mathrm{cf}}} \delta_{(X_i,S_i,Y_i)},
\]
where $N_{\mathrm{cf}}$ denotes the number of generated contrastive or counterfactual samples.

From a statistical perspective, these techniques enforce individual fairness by explicitly constraining the predictor to produce invariant or smoothly varying outputs across counterfactual pairs. This is typically operationalized through additional regularization terms or contrastive losses of the form
\[
\mathcal{L}_{\mathrm{cf}} = \sum_i d_P\big(f(X_i,S_i), f(X_i,S_i')\big),
\]
which penalize prediction discrepancies attributable solely to changes in the sensitive attribute.

Beyond individual-level consistency, augmentation-based methods can also promote GF by balancing representation across sensitive groups or by reducing the model’s reliance on sensitive attributes through exposure to symmetric counterfactual evidence. In this sense, counterfactual augmentation provides a unifying mechanism to address both fairness notions, with the trade-off controlled via the strength of the contrastive regularization or the generation strategy. Representative approaches such as \cite{calmon2017optimized, han2023dualfair, li2022accurate,lohia2019bias} illustrate how counterfactual or contrastive augmentation can simultaneously reduce group-level disparities while enforcing individual-level invariance.
    \end{itemize}

\paragraph{\textbf{Strategies based on regularization or optimization objectives with explicit trade-off control.}}
\begin{itemize}
      
\item \emph{Regularization}: These approaches incorporate fairness constraints directly into the model training objective. Rather than altering the data distribution or reweighting instances, they modify the learning dynamics by augmenting the loss function with terms that penalize unfair behavior at both the group and individual levels.

Formally, given a base predictive loss $L(f(X_i), Y_i)$, the typical in-pro\-cess\-ing objective takes the form
\[
\min_f \; \mathcal{L}(f)
  = \underbrace{\sum_i L(f(X_i), Y_i)}_{\text{prediction loss}}
  \;+\; \lambda_{\mathrm{GF}} \,\Phi_{\mathrm{group}}(f)
  \;+\; \lambda_{\mathrm{IF}} \,\Psi_{\mathrm{individual}}(f),
\]
where $\Phi_{\mathrm{group}}(f)$ penalizes statistical dependence between $S$ and either predictions $f(X)$ or errors. Common choices include demographic-parity-based moments, covariance constraints, or adversarial objectives. On the other hand, $\Psi_{\mathrm{individual}}(f)$ promotes local consistency, typically formalized via Lipschitz-type penalties, neighborhood consistency terms, or adversarial perturbation constraints, ensuring that similar inputs receive similar outputs.

From a statistical perspective, this corresponds to \textit{multi-objective optimization under fairness constraints}, where fairness regularizers act as shape restrictions on the hypothesis space. These constraints induce implicit smoothing or geometric structure in the learned predictor, with GF penalties reducing large-scale disparities across sensitive groups, while IF penalties encourage local smoothness of the decision boundary.

Regularization-based formulations naturally allow exploration of the \textit{Pareto frontier} between accuracy, group fairness, and individual fairness. By tuning $\lambda_{\mathrm{GF}}$ and $\lambda_{\mathrm{IF}}$, one can navigate trade-offs between global parity and local consistency, often revealing regimes where both improve jointly. Seminal contributions to this line of work include the approaches proposed by \cite{liu2025leveraging, han2023dualfair, li2022accurate, nagpal2025optimizing, wang2024individual2, wang2024individual, liu2024adversarial, long2023individual, dwork2012fairness}.

\item \emph{Metric-based approaches}: These approaches enforce fairness by explicitly penalizing discrepancies in predictions for pairs of individuals who are similar according to a predefined task-relevant metric. Instead of modifying the data distribution or introducing group-level invariance constraints, these methods embed geometric structure directly into the loss function, ensuring that the learned predictor behaves consistently within local neighborhoods of the input space.

Let $d_X(\cdot,\cdot)$ denote a similarity metric on the feature space. A typical formulation augments the prediction loss with a pairwise consistency penalty,
\[
\min_f \; \mathcal{L}(f)
  = \sum_i L(f(X_i), Y_i)
  \;+\; \lambda \sum_{i,j} \alpha_{ij}\,
      \bigl| f(X_i) - f(X_j) \bigr|,
\]
where $\alpha_{ij}$ is a weighting coefficient (often decreasing with $d_X(X_i,X_j)$) that emphasizes pairs of individuals who should be treated similarly. This structure enforces a form of Lipschitz continuity:
\[
\bigl| f(X_i) - f(X_j) \bigr| \leq C\, d_X(X_i,X_j),
\]
encouraging local smoothness of the decision function.

From a statistical and geometric standpoint, these methods correspond to \textit{regularization under similarity constraints}. The predictor is restricted to vary slowly in regions of high data density, thereby reducing individual-level inconsistencies and mitigating sensitivity to local perturbations.

Crucially, certain metric-based formulations can guarantee that individual fairness implies group fairness under appropriate conditions. For example, when sensitive groups occupy well-separated regions of the feature space or when group disparities emerge primarily from local inconsistencies, enforcing Lipschitz-like constraints can indirectly reduce global disparities across groups. Several works follow this line of reasoning by exploiting metric-based notions of similarity to connect local fairness constraints with group-level outcomes, including  \cite{li2022accurate, dwork2012fairness,small2024equalised} .

\item \emph{Adversarial learning}: These approaches introduce an adversarial component into the training pipeline to enforce fairness constraints through invariance. The predictor is trained to produce outputs or intermediate representations from which an adversary cannot recover the sensitive attribute $S$. Fairness emerges as a minimax optimization problem: the predictor aims to minimize prediction loss, while the adversary seeks to maximize their ability to infer $S$.

Formally, let $f$ denote the predictor and $g$ the adversary. A common formulation is:
\[
\min_f \; \max_g \;
\mathcal{L}(f,g)
=
\underbrace{\sum_i L(f(X_i), Y_i)}_{\text{prediction loss}}
\;-\;
\lambda_{\mathrm{adv}}
\underbrace{\sum_i L_{\mathrm{adv}}\bigl(g(f(X_i)), S_i\bigr)}_{\text{adversarial loss}},
\]
where $L_{\mathrm{adv}}$ encourages the adversary to accurately predict $S$, thereby pushing the predictor to remove information about $S$ from its outputs or latent representations. This form ensures that the learned representation is approximately invariant to the sensitive attribute, thereby promoting group fairness.

From a statistical perspective, this can be understood as enforcing \textit{conditional independence constraints} of the form
\[
f(X) \;\perp\!\!\!\perp\; S,
\]
achieved implicitly through adversarial pressure. The resulting model approximates a representation where the sensitive attribute carries minimal predictive signal.

To incorporate individual fairness, extensions introduce additional losses, such as:
\begin{itemize}
    \item \textbf{pairwise similarity penalties}, encouraging the predictor to preserve geometric or semantic distances between similar individuals;
    \item \textbf{counterfactual losses}, requiring predictions to remain stable under counterfactual modifications of sensitive attributes or non-sensitive perturbations.
    These additions ensure local consistency and robustness, aligning adversarial frameworks with IF principles.
\end{itemize}

Adversarial learning primarily advances \textit{group fairness} by removing $S$-related information from representations or predictions. When augmented with pairwise or counterfactual components, it also promotes \textit{individual fairness}, ensuring that local neighborhoods in the input space yield consistent outputs while maintaining invariance across groups. Some works adopt this perspective, including \cite{liu2024adversarial}.
\end{itemize} 

\paragraph{\textbf{Strategies based on local or post-hoc adjustments.}}
\begin{itemize} 
\item \emph{Thresholding}: Given a score-producing model $f(X)$, these methods introduce group-specific thresholds $\tau_s$ such that the resulting binary predictions
\[
\hat{Y} = \mathbb{I}\{f(X) \ge \tau_{S}\}
\]
align with desired group-fairness criteria, usually the \textit{equalized odds}.

From a statistical perspective, this corresponds to learning a post-processing mapping 
\[
g(f(X), S) \in \{0,1\}
\]
that minimizes misclassification error subject to constraints such as
\[
P(g(f(X), S)=1 \mid Y=y, S=s)
\approx
\]
\[
\approx
P(g(f(X), S)=1 \mid Y=y, S=s'),
\quad \forall y\in\{0,1\}.
\]

Different formulations extend these constraints in various directions. Thresholding-based post-processing primarily enforces \textit{group fairness} by equalizing error rates across sensitive groups. However, incorporating similarity metrics, local smoothness constraints, or instance-level stability terms enables partial alignment with \textit{individual fairness}, especially when the score distribution exhibits heterogeneous calibration across subpopulations.

Representative works within this line of approaches include \cite{small2024equalised}.
       
\item \emph{Reject-option}: These methods operate in an ``ambiguity region'' of the score space (typically around a decision boundary) where predictions are most uncertain and fairness violations are more likely to occur. Given a score model $f(X)$, the post-processor selectively modifies predictions for instances whose scores fall within an interval $[ \alpha, \beta ]$, assigning labels that reduce disparities across groups or correct local inconsistencies. Formally, the adjusted prediction can be written as
\[
\hat{Y}^{\text{adj}} =
\begin{cases}
1, & f(X) > \beta, \\
0, & f(X) < \alpha, \\
h(f(X), S), & \alpha \le f(X) \le \beta,
\end{cases}
\] 
where $h(\cdot)$ is a fairness-driven correction rule.

From a statistical perspective, reject-option mechanisms can be interpreted as local recalibrations of the decision rule, targeting regions where the conditional distributions $P(Y \mid f(X), S=s)$ differ most across groups. The ambiguity interval allows the post-processor to enforce constraints such as
\[
\operatorname{sign}(f(X_i) - f(X_j)) \approx 
\operatorname{sign}(\hat{Y}_i^{\text{adj}} - \hat{Y}_j^{\text{adj}})
\]
for similar individuals $(i,j)$, thereby simultaneously addressing group-level disparities and individual-level inconsistencies.

Reject-option post-processing typically improves \textit{group fairness} by strategically altering decisions near the boundary where misclassification disparities are concentrated. Moreover, because adjustments are made at the instance level, these methods naturally support \textit{individual fairness}, especially when the correction rule leverages similarity structure or smoothness constraints.
Representative examples include \cite{lohia2019bias}.
\end{itemize} 

\begin{table}[htbp]
\centering
\scriptsize
\rotatebox{0}{
\begin{tabularx}{\textwidth}{l l p{1cm} p{2.8cm} X X}
\hline
\textbf{Ref.} & \textbf{Year} & \textbf{Stage} & \textbf{Sub-taxonomy} & \textbf{Notes} & \textbf{Code} \\
\hline
\cite{arnaiz2023towards} & 2023 & PRE & Reweighting and resampling & IF+GF via Reweighting based on Shapley values & \url{https://github.com/youlei202/FairSHAP} \\
\cite{calmon2017optimized} & 2017 & PRE & Data transformation + Augmentation and Contrastive & Optimized pre-processing (soft version of IF) & \url{https://github.com/fair-preprocessing/nips2017} \\
\cite{zemel2013} & 2013 & PRE & Representation and embedding & Fair representations & \url{https://github.com/zjelveh/learning-fair-representations} \\
\cite{xu2024compatibility} & 2024 & PRE / POST & Data transformation & Stablishes theoretical guarantees to obtain GF+IF & \url{https://github.com/xushizhou/Compatibility-Group-Individual-Fairness}  \\
\cite{liu2025leveraging} & 2025 & IN & Regularization + Representation and embedding &   Fair representation + multi-objective & Based on \url{https://github.com/HobbitLong/SupContrast.git.}\\
\cite{nagpal2025optimizing} & 2025 & IN & Regularization & Multi-objective: IF+GF & \xmark \\
\cite{wang2024individual2} & 2024 & IN & Regularization  & Multi-objective: IF+GF & \xmark \\
\cite{wang2024individual} & 2024 & IN & Regularization  & Multi-objective: IF+GF & \xmark \\
\cite{liu2024adversarial} & 2024 & IN & Adversarial learning + Regularization & Loss with IF, GF and classification terms & \url{https://github.com/satansin/GIFair} \\
\cite{han2023dualfair} & 2023 & IN & Augmentation  and Contrastive + Representation and embedding + Regularization & IF (counterfactual) + GF & \url{https://github.com/Sungwon-Han/DualFair} \\
\cite{long2023individual} & 2023 & IN   & Regularization  & Model ensembles reduce individual arbitrariness & \xmark \\
\cite{li2022accurate} & 2022 & IN & Metric-based approaches +  Regularization + Augmentation and Contrastive & IF $\implies$ GF (Stat Parity) via new metrics & \url{https://github.com/Xuran-LI/AccurateFairnessCriterion} \\
\cite{dwork2012fairness} & 2012 & IN & Metric-based approaches + Regularization & First IF paper; optimization framework & \url{https://github.com/dodger487/fairness} \\
\cite{small2024equalised} & 2024 & POST & Thresholding + Metric-based approaches & Equalized odds + similarity-based IF & \xmark \\
\cite{lohia2019bias} & 2019 & POST & Augmentation and  Contrastive + Reject-option & Reject Option variant ensuring IF+GF & \xmark \\
\cite{binns2020apparent} & 2020 & - & Theoretical / Conceptual & Seminal paper on GF–IF trade-off (philosophical) & \xmark \\

\hline
\end{tabularx}
}
\caption{Reduced summary of selected works addressing both group and individual fairness, reordered by column.}
\label{tab:table_trade-offs}
\end{table}

\subsection{Benchmarking, implementation and method statistics}
\label{sec:method_stats}

In the previous section, we introduced a taxonomy of methods for managing trade-offs between GF and IF and reviewed representative approaches within each category. We now shift attention from theoretical formulation and taxonomy to the practical considerations that govern the implementation and evaluation of these methods. Beyond their formulation, the feasibility of jointly enforcing GF and IF depends on factors including computational complexity, benchmark datasets, and experimental design choices.

This section analyzes these implementation aspects within the surveyed literature. We examine how methods are distributed across different stages of the machine learning pipeline, the types of datasets and evaluation protocols they rely on, and the computational assumptions they make. Table~\ref{tab:table_trade-offs} and Table~\ref{tab:table_datasets} summarize these characteristics, providing a consolidated view of how trade-off-aware fairness methods are operationalized in empirical studies.

\begin{table}[htbp]
\centering
\scriptsize
\begin{tabularx}{\textwidth}{l l l p{3cm} X X}
\hline
\textbf{Ref.} & \textbf{Year} & \textbf{Stage} & \textbf{Sub-taxonomy} & \textbf{Computational cost} & \textbf{Datasets} \\
\hline
\cite{arnaiz2023towards} & 2023 & PRE & Reweighting and resampling & Model dependent: Tied to base learner. FairShap is competitive up to $n=30k$. Benchmarks: 0.03s to 453s. & COMPAS, Adult, German Credit, FairFace, LFWA, CelebA \\ 
\cite{calmon2017optimized} & 2017 & PRE & Data transformation + Augmentation and Contrastive & Feature-dependent: High efficiency for low-dimensional spaces via CVXPY. & Adult, COMPAS \\ 
\cite{zemel2013} & 2013 & PRE & Representation and embedding & High efficiency: Utilizes Numba (JIT compiler). Ranges from seconds to minutes; scales poorly with high $n$. & Adult, German credit, HHP \\ 
\cite{xu2024compatibility} & 2024 & PRE/POST & Data transformation & Variable: Performance varies by test; $>1$ hour for LSAC vs. $<2$ minutes for CRIME. & CRIME, LSAC \\ 
\cite{liu2025leveraging} & 2025& IN & Regularization + Representation and embedding & High cost: Explicitly noted as demanding; potentially inaccessible for limited budgets. & OULAD \\ 
\cite{nagpal2025optimizing} & 2025 & IN & Regularization & - & Adult, German Credit \\ 
\cite{wang2024individual2} & 2024 & IN & Regularization & - & COMPAS, ROSSI, KKBox, Support \\ 
\cite{wang2024individual} & 2024 & IN & Regularization & - & German Credit, CreditRisk, Facebook, Bail \\ 
\cite{liu2024adversarial} & 2024 & IN & Adversarial learning + Regularization & Moderate: Benchmarks estimate 30–45 min for Adult and 45–60 min for COMPAS. & Adult, German Credit, COMPAS \\ 
\cite{han2023dualfair} & 2023 & IN & Augmentation and Contrastive + Representation and embedding + Regularization & High per-sample cost: Driven by C-VAE and pairwise comparisons. It can take several hours. & Adult, German Credit, COMPAS, LSAC, Students, Communities \\ 
\cite{long2023individual} & 2023 & IN & Regularization & Expensive: Requires iterative re-training of multiple models. & Adult, HSLS, ENEM \\ 
\cite{li2022accurate} & 2022 & IN & Metric-based approaches + Regularization + Augmentation and Contrastive & Optimized: Augmentation uses only extremal values to minimize consumption. & Adult, German Credit, COMPAS, Ctrip \\ 
\cite{dwork2012fairness} & 2012 & IN & Metric-based approaches + Regularization & - & - \\ 
\cite{small2024equalised} & 2024 & POST & Thresholding + Metric-based approaches & Low cost: Utilizes continuous functions described as "easy to compute." & COMPAS, CreditRisk \\ 
\cite{lohia2019bias} & 2019 & POST & Augmentation and Contrastive + Reject-option & Expensive: Intensive despite internal algorithmic efficiencies. & Adult, German Credit, COMPAS \\ 
\cite{binns2020apparent} & 2020 & - & Theoretical / Conceptual & - & German Credit \\
\hline
\end{tabularx}
\caption{Reduced summary of selected works addressing both group and individual fairness, including computational cost and datasets used.}
\label{tab:table_datasets}
\end{table}

\begin{figure}[ht!]
    \centering
    \includegraphics[width=\linewidth]{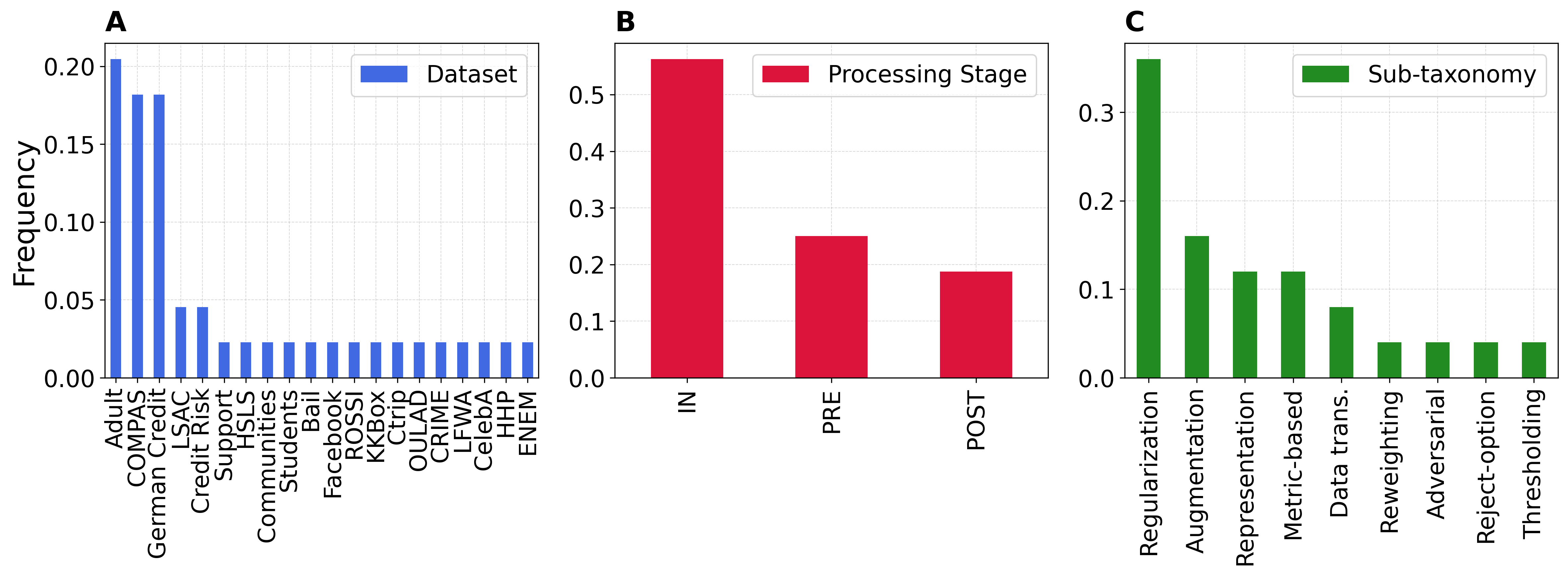}
    \caption{\textbf{Summary statistics of papers addressing the GF vs IF trade-off.} (A) Dataset distribution, showing heavy reliance on a few benchmark datasets (e.g., Adult, COMPAS).  (B) Distribution of methods by processing stage, with in-processing approaches being the most common. (C) Proportion of papers within each sub-taxonomy, where regularization is the most common approach.}
    \label{fig:trade-off_stats}
\end{figure}

As illustrated in Table~\ref{tab:table_datasets} and Figure~\ref{fig:trade-off_stats}~(A), most trade-off methods are evaluated on the \emph{Adult}, \emph{COMPAS}, and \emph{German Credit} datasets, each appearing in approximately 20\% of the surveyed studies. Although these widely used benchmarks enable comparisons across methods, their limited variety raises concerns about the extent to which current findings generalize to higher-dimensional domains, such as Computer Vision or Natural Language Processing, which remain largely underrepresented in the literature.

Across the surveyed literature, computational efficiency is rarely treated as a first-order concern. Despite the increasing complexity of hybrid models, only one study provides a systematic analysis of the trade-offs between fairness objectives, predictive accuracy, and computational cost \cite{arnaiz2023towards}. Research specifically addressing the "cost of auditing" or the "cost of training" for joint fairness is remarkably scarce. Several papers acknowledge these issues only qualitatively. For instance, \cite{liu2025leveraging} acknowledges that the demand for intense computational power can make their method inaccessible to practitioners with limited budgets. Similarly, \cite{long2023individual} notes that, although their technique achieves fairness with minimal accuracy loss, it is computationally expensive because it requires retraining multiple models. Even when efficiency measures are explicitly enacted, algorithms such as the one proposed by \cite{lohia2019bias} remain computationally demanding. Other works, such as \cite{li2022accurate}, report model modifications specifically aimed at reducing computational costs but do not provide a comparative study of how these costs scale with varying fairness constraints. The lack of systematic analysis of computational times highlights an important gap in the literature and underscores the need for principled frameworks that jointly account for fairness, performance, and computational efficiency.

Figure~\ref{fig:trade-off_stats}~(B) details the distribution of methods across the processing pipeline. \textit{In-processing} methods are the most prevalent category (60\%), reflecting a preference for integrating fairness directly into the optimization landscape. However, pre-processing (25\%) and post-processing (19\%) methods maintain a balanced presence, highlighting the continued need for model-agnostic solutions that do not require altering their architecture.

Regarding resource availability, Table~\ref{tab:table_trade-offs} shows that only 56\% of the reviewed methods provide publicly accessible code. The absence of open implementations for nearly half of the surveyed approaches limits independent verification of the reported trade-offs and poses practical barriers to their adoption in real-world and industrial settings.

\section{Discussion}
\label{sec:discusion}

In this paper, we surveyed machine learning methods that explicitly address the trade-offs between GF and IF, two central yet often conflicting notions in the algorithmic fairness literature. While theoretical research shows that GF and IF are mathematically incompatible except in trivial cases \cite{xu2024compatibility, zhou2022group}, and that similar impossibility results arise even within group-based notions \cite{gao2025fair}, a growing body of work proposes methods that explicitly manage the resulting trade-offs. By focusing on approaches that jointly consider both perspectives, we highlighted how fairness trade-offs are operationalized in practice, the assumptions on which they rely, and the consequences they entail. Rather than focusing on a single definition of fairness, the reviewed works reflect a growing recognition that fairness is inherently multi-dimensional and that managing tensions between competing criteria is unavoidable in real-world systems \cite{bernard2025systematic}.

We introduce a taxonomy of the surveyed methods, encompassing flexible mechanisms such as in-processing regularization, multi-objective optimization, and constrained learning. These approaches allow practitioners to navigate competing fairness objectives in a controlled and interpretable manner. By making explicit the costs associated with prioritizing one notion of fairness over another, they support more informed and accountable design choices.

Moreover, our review reveals that the field remains at an early stage of maturity. Most methods are evaluated on a narrow set of benchmark datasets, with limited exploration of high-dimensional or domain-specific settings. Empirical evaluations often emphasize fairness and accuracy, while practical considerations such as computational efficiency, scalability, and reproducibility receive comparatively little attention. In addition, although many approaches offer tunable trade-offs, there is limited guidance on how to select these parameters in practice or align them with legal, institutional, or societal constraints \cite{papagiannidis2025responsible, bach2025insights}.

These observations point to several promising directions for future research. First, there is a need for more systematic evaluation protocols that assess fairness trade-offs alongside computational cost and deployment constraints. Second, expanding empirical studies to more diverse application domains would improve understanding of how joint fairness methods behave beyond standard tabular benchmarks. Third, closer integration with causal, legal, and human-centered perspectives could help ground trade-off decisions in contextual and normative considerations, rather than treating them solely as technical optimization problems. Finally, a significant practical barrier is that most approaches assume access to fully labeled sensitive attributes during training, a strong assumption that is rarely met in real-world applications or regulated environments. A few exceptions include \cite{wang2024individual2}, which explicitly considers censored datasets.

Additionally, the surveyed literature consistently points toward the need for more flexible and expressive fairness frameworks that move beyond rigid parity constraints. Several studies emphasize enforcing fairness within learned representations rather than solely through aggregate metrics, suggesting that richer latent spaces can better support joint group- and individual-level guarantees \cite{zemel2013, han2023dualfair}. Reweighting pre-processing approaches, such as FairShap, further highlight the potential of fairness tools not only for evaluation, but also for generating counterfactual explanations and informing data reweighting or generation strategies \cite{arnaiz2023towards}.

Conversely, some recent works challenge the inevitability of the conflict itself, proposing notions like `accurate fairness' where ensuring individual treatment aligns with ground truth accuracy, thereby suggesting that trade-offs may sometimes stem from data bias rather than mathematical incompatibility \cite{li2022accurate}.

Furthermore, many studies highlight the gap between fairness objectives and practical deployment, including limited access to model internals, high computational costs, and predictive multiplicity, which motivate post-processing and black-box methods that remain effective under practical constraints \cite{lohia2019bias, long2023individual}. Recent extensions to intersectional, partially observed, and networked data underscore the need for fairness methods that operate across a broader range of datasets, common in realistic settings \cite{liu2025leveraging, wang2024individual, wang2024individual2}. Finally, several authors argue that apparent conflicts between GF and IF may stem from how fairness metrics are applied to data shaped by existing social inequities, rather than from irreconcilable normative principles \cite{binns2020apparent}.

Methods that explicitly manage trade-offs between GF and IF represent an important step toward more responsible and transparent algorithmic decision-making. By clarifying the structure of existing approaches and identifying key gaps, this survey aims to provide a foundation for future work that advances both the theoretical rigor and practical relevance of hybrid fairness methods.

\section*{Acknowledgements}
This paper has been carried out within the framework of the Recovery, Transformation and Resilience Plan funds, financed by the European Union (Next Generation) through the grant ``Cátedras ENIA 2022 para la creación de cátedras universidad-empresa en IA'' AImpulsa: Cátedra UC3M-Universia de Economía del Dato y la Inteligencia Artificial Responsable aplicada a la Creación Exponencial de Valor. It has been also partially supported by the Ministerio de Ciencia e Innovación, Gobierno de España, (grants PID2022-137243OB-I00 and PID2022-137818OB-I00 funded by MCIN/AEI/10.13039/501100011033 and European Union NextGenerationEU/PRTR).

\bibliographystyle{unsrt}
\bibliography{refs}

\begin{thebibliography}{10}

\bibitem{chiang2022exploring}
Chun-Wei Chiang and Ming Yin.
\newblock Exploring the effects of machine learning literacy interventions on laypeople’s reliance on machine learning models.
\newblock In {\em Proceedings of the 27th International Conference on Intelligent User Interfaces}, pages 148--161, 2022.

\bibitem{barocas2023fairness}
Solon Barocas, Moritz Hardt, and Arvind Narayanan.
\newblock {\em Fairness and machine learning: Limitations and opportunities}.
\newblock MIT press, 2023.

\bibitem{papagiannidis2025responsible}
Emmanouil Papagiannidis, Patrick Mikalef, and Kieran Conboy.
\newblock Responsible artificial intelligence governance: A review and research framework.
\newblock {\em The Journal of Strategic Information Systems}, 34(2):101885, 2025.

\bibitem{laufer2025constitutes}
Benjamin Laufer, Manish Raghavan, and Solon Barocas.
\newblock What constitutes a less discriminatory algorithm?
\newblock In {\em Proceedings of the 2025 Symposium on Computer Science and Law}, pages 136--151, 2025.

\bibitem{mehrabi2021survey}
Ninareh Mehrabi, Fred Morstatter, Nripsuta Saxena, Kristina Lerman, and Aram Galstyan.
\newblock A survey on bias and fairness in machine learning.
\newblock {\em ACM computing surveys (CSUR)}, 54(6):1--35, 2021.

\bibitem{hort2024bias}
Max Hort, Zhenpeng Chen, Jie~M Zhang, Mark Harman, and Federica Sarro.
\newblock Bias mitigation for machine learning classifiers: A comprehensive survey.
\newblock {\em ACM Journal on Responsible Computing}, 1(2):1--52, 2024.

\bibitem{dwork2012fairness}
Cynthia Dwork, Moritz Hardt, Toniann Pitassi, Omer Reingold, and Richard Zemel.
\newblock Fairness through awareness.
\newblock In {\em Proceedings of the 3rd innovations in theoretical computer science conference}, pages 214--226, 2012.

\bibitem{bernard2025systematic}
Nolwenn Bernard and Krisztian Balog.
\newblock A systematic review of fairness, accountability, transparency, and ethics in information retrieval.
\newblock {\em ACM Computing Surveys}, 57(6):1--29, 2025.

\bibitem{anderson2025algorithmic}
Joshua~W Anderson and Shyam Visweswaran.
\newblock Algorithmic individual fairness and healthcare: a scoping review.
\newblock {\em JAMIA open}, 8(1):ooae149, 2025.

\bibitem{xu2024compatibility}
Shizhou Xu and Thomas Strohmer.
\newblock On the (in) compatibility between group fairness and individual fairness.
\newblock {\em arXiv preprint arXiv:2401.07174}, 2024.

\bibitem{zhou2022group}
Wanying Zhou.
\newblock {\em Group vs. individual algorithmic fairness}.
\newblock PhD thesis, University of Southampton, 2022.

\bibitem{caton2024survey}
Simon Caton and Christian Haas.
\newblock Fairness in machine learning: A survey.
\newblock {\em ACM computing surveys (CSUR)}, 56(7), 2024.

\bibitem{gao2025fair}
Jianhui Gao, Benson Chou, Zachary~R McCaw, Hilary Thurston, Paul Varghese, Chuan Hong, and Jessica Gronsbell.
\newblock What is fair? defining fairness in machine learning for health.
\newblock {\em Statistics in Medicine}, 44(20-22):e70234, 2025.

\bibitem{rabonato2025systematic}
Ricardo~Trainotti Rabonato and Lilian Berton.
\newblock A systematic review of fairness in machine learning.
\newblock {\em AI and Ethics}, 5(3):1943--1954, 2025.

\bibitem{pagano2023bias}
Tiago~P Pagano, Rafael~B Loureiro, Fernanda~VN Lisboa, Rodrigo~M Peixoto, Guilherme~AS Guimar{\~a}es, Gustavo~OR Cruz, Maira~M Araujo, Lucas~L Santos, Marco~AS Cruz, Ewerton~LS Oliveira, et~al.
\newblock Bias and unfairness in machine learning models: a systematic review on datasets, tools, fairness metrics, and identification and mitigation methods.
\newblock {\em Big data and cognitive computing}, 7(1):15, 2023.

\bibitem{alves2023survey}
Guilherme Alves, Fabien Bernier, Miguel Couceiro, Karima Makhlouf, Catuscia Palamidessi, and Sami Zhioua.
\newblock Survey on fairness notions and related tensions.
\newblock {\em EURO journal on decision processes}, 11:100033, 2023.

\bibitem{kozodoi2022fairness}
Nikita Kozodoi, Johannes Jacob, and Stefan Lessmann.
\newblock Fairness in credit scoring: Assessment, implementation and profit implications.
\newblock {\em European Journal of Operational Research}, 297(3):1083--1094, 2022.

\bibitem{calders2010three}
Toon Calders and Sicco Verwer.
\newblock Three naive bayes approaches for discrimination-free classification.
\newblock {\em Data mining and knowledge discovery}, 21(2):277--292, 2010.

\bibitem{hardt2016equality}
Moritz Hardt, Eric Price, and Nati Srebro.
\newblock Equality of opportunity in supervised learning.
\newblock {\em Advances in neural information processing systems}, 29, 2016.

\bibitem{ramos2022robust}
Guilherme Ramos, Ludovico Boratto, and Mirko Marras.
\newblock Robust reputation independence in ranking systems for multiple sensitive attributes.
\newblock {\em Machine Learning}, 111(10):3769--3796, 2022.

\bibitem{zehlike2020reducing}
Meike Zehlike and Carlos Castillo.
\newblock Reducing disparate exposure in ranking: A learning to rank approach.
\newblock In {\em Proceedings of the web conference 2020}, pages 2849--2855, 2020.

\bibitem{friedler2021possibility}
Sorelle~A Friedler, Carlos Scheidegger, and Suresh Venkatasubramanian.
\newblock The (im) possibility of fairness: Different value systems require different mechanisms for fair decision making.
\newblock {\em Communications of the ACM}, 64(4):136--143, 2021.

\bibitem{han2023dualfair}
Sungwon Han, Seungeon Lee, Fangzhao Wu, Sundong Kim, Chuhan Wu, Xiting Wang, Xing Xie, and Meeyoung Cha.
\newblock Dualfair: fair representation learning at both group and individual levels via contrastive self-supervision.
\newblock In {\em Proceedings of the ACM web conference 2023}, pages 3766--3774, 2023.

\bibitem{lohia2019bias}
Pranay~K Lohia, Karthikeyan~Natesan Ramamurthy, Manish Bhide, Diptikalyan Saha, Kush~R Varshney, and Ruchir Puri.
\newblock Bias mitigation post-processing for individual and group fairness.
\newblock In {\em Icassp 2019-2019 ieee international conference on acoustics, speech and signal processing (icassp)}, pages 2847--2851. IEEE, 2019.

\bibitem{lahoti2019ifair}
Preethi Lahoti, Krishna~P Gummadi, and Gerhard Weikum.
\newblock ifair: Learning individually fair data representations for algorithmic decision making.
\newblock In {\em 2019 ieee 35th international conference on data engineering (icde)}, pages 1334--1345. IEEE, 2019.

\bibitem{pessach2022review}
Dana Pessach and Erez Shmueli.
\newblock A review on fairness in machine learning.
\newblock {\em ACM Computing Surveys (CSUR)}, 55(3):1--44, 2022.

\bibitem{ferrara2024fairness}
Emilio Ferrara.
\newblock Fairness and bias in artificial intelligence: A brief survey of sources, impacts, and mitigation strategies.
\newblock {\em Sci}, 6(1):3, 2024.

\bibitem{shah2025comprehensive}
Milind Shah and Nitesh Sureja.
\newblock A comprehensive review of bias in deep learning models: Methods, impacts, and future directions.
\newblock {\em Archives of Computational Methods in Engineering}, 32(1):255--267, 2025.

\bibitem{joseph2016fairness}
Matthew Joseph, Michael Kearns, Jamie~H Morgenstern, and Aaron Roth.
\newblock Fairness in learning: Classic and contextual bandits.
\newblock {\em Advances in neural information processing systems}, 29, 2016.

\bibitem{corbett2023measure}
Sam Corbett-Davies, Johann~D Gaebler, Hamed Nilforoshan, Ravi Shroff, and Sharad Goel.
\newblock The measure and mismeasure of fairness.
\newblock {\em Journal of Machine Learning Research}, 24(312):1--117, 2023.

\bibitem{calmon2017optimized}
Flavio Calmon, Dennis Wei, Bhanukiran Vinzamuri, Karthikeyan Natesan~Ramamurthy, and Kush~R Varshney.
\newblock Optimized pre-processing for discrimination prevention.
\newblock {\em Advances in neural information processing systems}, 30, 2017.

\bibitem{arnaiz2023towards}
Adrian Arnaiz-Rodriguez and Nuria Oliver.
\newblock Towards algorithmic fairness by means of instance-level data re-weighting based on shapley values.
\newblock {\em arXiv preprint arXiv:2303.01928}, 2023.

\bibitem{zemel2013}
Rich Zemel, Yu~Wu, Kevin Swersky, Toniann Pitassi, and Cynthia Dwork.
\newblock Learning fair representations.
\newblock In {\em International Conference on Machine Learning}, 2013.

\bibitem{liu2025leveraging}
Zifeng Liu, Wanli Xing, Yihan Jiang, Chenglu Li, Taehyun Kim, and Hai Li.
\newblock Leveraging contrastive learning to improve group and individual fairness in predictive analytics for online learning.
\newblock {\em Journal of Computing in Higher Education}, pages 1--30, 2025.

\bibitem{li2022accurate}
Xuran Li, Peng Wu, and Jing Su.
\newblock Accurate fairness: Improving individual fairness without trading accuracy.
\newblock In {\em Proceedings of the AAAI Conference on Artificial Intelligence}, volume~37, pages 14312--14320, 2022.

\bibitem{nagpal2025optimizing}
Rashmi Nagpal, Rasoul Shahsavarifar, Vaibhav Goyal, and Amar Gupta.
\newblock Optimizing fairness and accuracy: a pareto optimal approach for decision-making.
\newblock {\em AI and Ethics}, 5(2):1743--1756, 2025.

\bibitem{wang2024individual2}
Zichong Wang, Jocelyn Dzuong, Xiaoyong Yuan, Zhong Chen, Yanzhao Wu, Xin Yao, and Wenbin Zhang.
\newblock Individual fairness with group awareness under uncertainty.
\newblock In {\em Joint European Conference on Machine Learning and Knowledge Discovery in Databases}, pages 89--106. Springer, 2024.

\bibitem{wang2024individual}
Zichong Wang, David Ulloa, Tongjia Yu, Raju Rangaswami, Roland Yap, and Wenbin Zhang.
\newblock Individual fairness with group constraints in graph neural networks.
\newblock In {\em Proceedings of the 27th European Conference on Artificial Intelligence (ECAI), 2023}. IOS Press, 2024.

\bibitem{liu2024adversarial}
Hao Liu and Raymond Chi-Wing Wong.
\newblock Adversarial learning of group and individual fair representations.
\newblock In {\em Pacific-Asia Conference on Knowledge Discovery and Data Mining}, pages 181--193. Springer, 2024.

\bibitem{long2023individual}
Carol Long, Hsiang Hsu, Wael Alghamdi, and Flavio Calmon.
\newblock Individual arbitrariness and group fairness.
\newblock {\em Advances in Neural Information Processing Systems}, 36:68602--68624, 2023.

\bibitem{small2024equalised}
Edward Small, Kacper Sokol, Daniel Manning, Flora~D Salim, and Jeffrey Chan.
\newblock Equalised odds is not equal individual odds: Post-processing for group and individual fairness.
\newblock In {\em Proceedings of the 2024 ACM conference on fairness, accountability, and transparency}, pages 1559--1578, 2024.

\bibitem{binns2020apparent}
Reuben Binns.
\newblock On the apparent conflict between individual and group fairness.
\newblock In {\em Proceedings of the 2020 conference on fairness, accountability, and transparency}, pages 514--524, 2020.

\bibitem{bach2025insights}
Tita~Alissa Bach, Magnhild Kaarstad, Elizabeth Solberg, and Aleksandar Babic.
\newblock Insights into suggested responsible ai (rai) practices in real-world settings: a systematic literature review.
\newblock {\em AI and Ethics}, pages 1--48, 2025.

\end{thebibliography}

\end{document}